\documentclass[letterpaper, 10 pt, conference]{ieeeconf}

\IEEEoverridecommandlockouts                              

\usepackage{amsmath,amssymb,amsfonts}
\usepackage{multirow}
\usepackage{graphicx}
\usepackage{rotating}
\usepackage{multicol}
\usepackage{amsmath}
\usepackage{float}
\usepackage{array}
\usepackage{caption}
\usepackage{subcaption}
\usepackage{siunitx}
\usepackage{float}
\usepackage{xcolor}
\usepackage{multirow}
\usepackage{graphicx}
\usepackage{color, soul, colortbl}
\usepackage{nicematrix}
\usepackage{colortbl} 
\usepackage{hyperref}

\DeclareUnicodeCharacter{2061}{}

\begin{document}

\title{\LARGE Marine$\mathcal{X}$: Design and Implementation of Unmanned Surface Vessel for Vision Guided Navigation}

\author{Muhayy Ud Din, Ahmed Humais, Waseem Akram, Mohamed Alblooshi , Lyes Saad Saoud, \\ Abdelrahman Alblooshi, Lakmal Seneviratne, and Irfan Hussain$^{*}$                                                     
\thanks{Khalifa University Center for Autonomous Robotic Systems (KUCARS), Khalifa University, United Arab Emirates.}%
\thanks{$^{*}$ Corresponding Author, Email: irfan.hussain@ku.ac.ae}
}
\maketitle

\begin{abstract}
Marine robots, particularly Unmanned Surface Vessels (USVs), have gained considerable attention for their diverse applications in maritime tasks, including search and rescue, environmental monitoring, and maritime security. This paper presents the design and implementation of a USV named marine$\mathcal{X}$. The hardware components of marine$\mathcal{X}$ are meticulously developed to ensure robustness, efficiency, and adaptability to varying environmental conditions.
Furthermore, the integration of a vision-based object tracking algorithm empowers marine$\mathcal{X}$ to autonomously track and monitor specific objects on the water surface. The control system utilizes PID control, enabling precise navigation of marine$\mathcal{X}$ while maintaining a desired course and distance to the target object.
To assess the performance of marine$\mathcal{X}$, comprehensive testing is conducted, encompassing simulation, trials in the marine pool, and real-world tests in the open sea. The successful outcomes of these tests demonstrate the USV's capabilities in achieving real-time object tracking, showcasing its potential for various applications in maritime operations.
\end{abstract}



\IEEEpeerreviewmaketitle

\section{Introduction}
 Marine robotics emerging as a significant field to explore and understand the complex marine ecosystem.
 There are two main classes of marine robots that are underwater vehicles and surface vehicles. The underwater vehicles are mainly used for underwater resource exploration, inspection, studying marine life, and for the collection of crucial data to study climate change. The surface vehicles, such as Unmanned surface vessels (USVs) are playing a significant role to enhance the maritime security~\cite{johnston2017marine}, perform the search and rescue operations~\cite{matos2013development}, environmental monitoring~\cite{Villa2016}, and for disaster response~\cite{Jorge2019}. The use of USVs in these areas enhances efficiency and minimizes the risks to human life.

Unmanned surface vehicles are complex in design as they required consistent coordination among their mechanical components, sensors, actuators, and navigation algorithms to effectively carry out their designated tasks. 
The mechanical configuration of a USV is identical to a conventional vessel, however, smaller in size~\cite{al2019design}. 
There are various types of vessel design among them, mono-hull design and catamaran (dual hull) design are most commonly used~\cite{papanikolaou2014ship}. 
The catamaran design provides enhanced stability and payload capacity compared to mono hull designs, which makes it a popular choice for the designing of testing platforms in the research field~\cite{caccia2006autonomous}~\cite{liu2016unmanned}. Despite the mechanical design, it is also required to build the Guidance, Navigation, and Control system~\cite{fossen2011handbook}. Furthermore, path planning, collision avoidance~\cite{mei2017smart} and path following~\cite{breivik2004path} are essential for USV to perform the navigation task autonomously. 

\begin{figure}
    \includegraphics[width=\columnwidth]{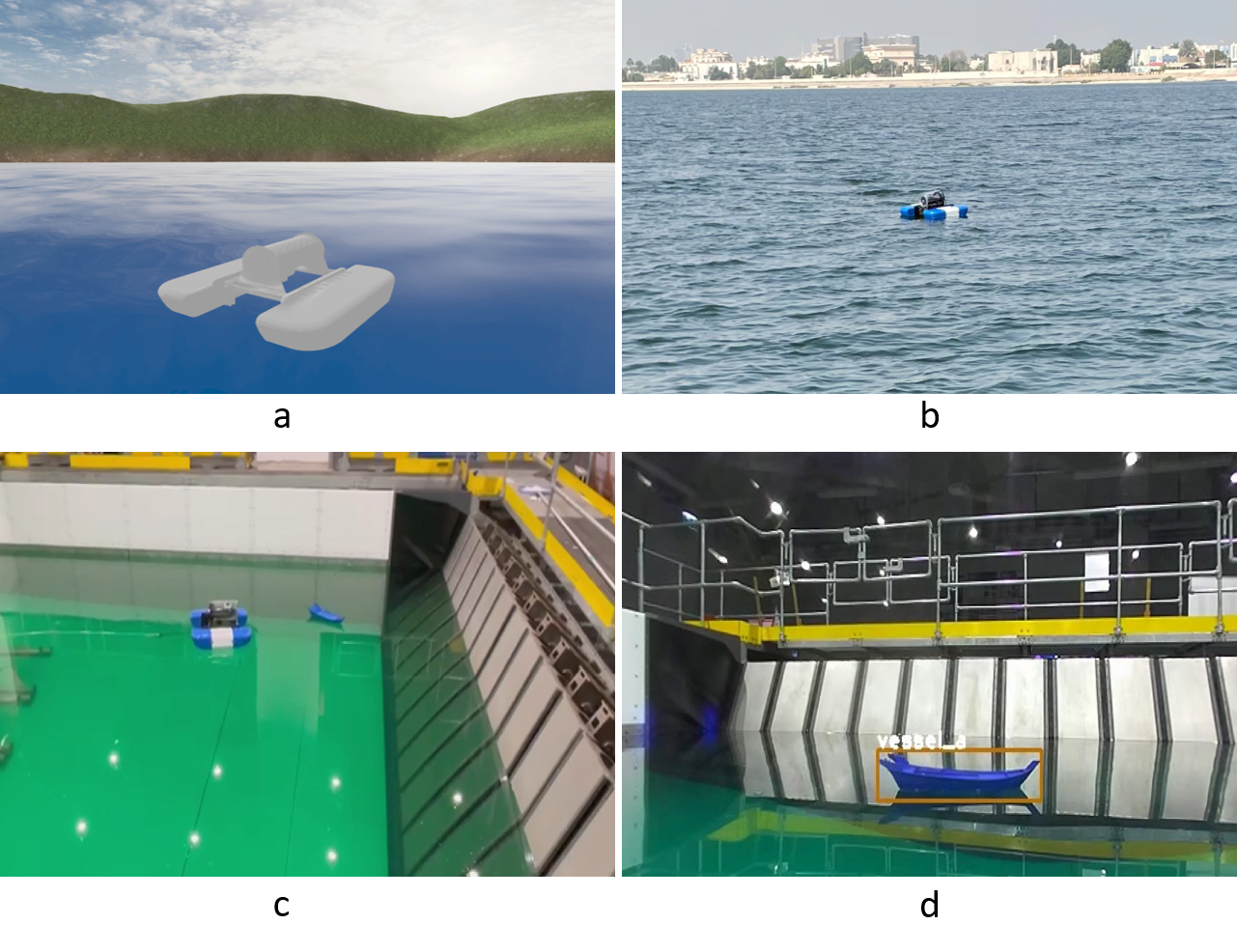}
    \caption{Marine$\mathcal{X}$ operating in different environments. a) represents Gazebo simulation, b) represents real sea experiment, c) shows the experiment in the marine pool, and d) shows the output of marine$\mathcal{X}$ onboard camera image after applying the object detection algorithm.}
    \label{fig:frontimg}
\end{figure}

In recent years, several USVs are developed by the research community. These systems are designed by keeping in mind the specific application to be performed. For instance, SESAMO~\cite{caccia2005autonomous} is developed with the catamaran design for the sampling of sea-surface microlayer. ROAV~\cite{ferreira2007roaz} is a USV, designed for river applications, with the capability of carrying up to a 50kg payload.  In order to demonstrate autonomous navigation capabilities, ARTMEMIS~\cite{manley2000evolution} was developed by MIT. Several commercial USVs are also developed by robotic companies, such as,  Wave Glider ~\cite{manley2010wave}, it is a mono-hull design USV developed by Liquid Robotics 
Inc (\url{https://www.liquid-robotics.com/}). Otter is a catamaran design developed by Maritime Robotics (\url{https://www.maritimerobotics.com/otter}). The commercial USVs are well equipped with functionalities However, they are very expensive. This leads us to the motivation of developing our own USV for particular tracking applications.

Several control algorithms such as Proportional Integral Derivative (PID) control, Model predictive control, and neural network-based control are proposed to navigate the USV autonomously~\cite{azzeri2015review}. In order to estimate the position of the USV, all these control algorithms required localization strategies such as GPS-based localization. Visual servo control provides real-time visual feedback to guide the motion and control of a robot, enabling precise and dynamic interactions with the environment based on visual information. It does not require explicit localization and it is widely applied in tasks such as object tracking and autonomous navigation~\cite{ahmed2023visionbased}. 

This study presents the design and fabrication of unmanned surface vessels. Moreover, a vision-guided navigation algorithm is developed that autonomously navigates marine$\mathcal{X}$ toward the given target. 

\textit{Contributions:} This work makes significant contributions in the following aspects:

\begin{itemize}
\item Design and Development: We successfully designed and developed the unmanned surface vessel, marine$\mathcal{X}$, using the robust catamaran design. This choice ensures adaptability to diverse environmental conditions, making it ideal for various maritime tasks.

\item Vision-Based Feedback Control: The implementation of a vision-based feedback control system using PID control allows precise visual servoing and efficient object tracking on the water surface.

\item 
Comprehensive Testing: Extensive testing, including simulations, indoor marine pool evaluations, and real-world open sea scenarios, validates marine$\mathcal{X}$'s design and autonomous navigation algorithm. The results highlight its capabilities for diverse maritime operations.
\end{itemize}




\begin{figure}
    \includegraphics[width=0.95\columnwidth]{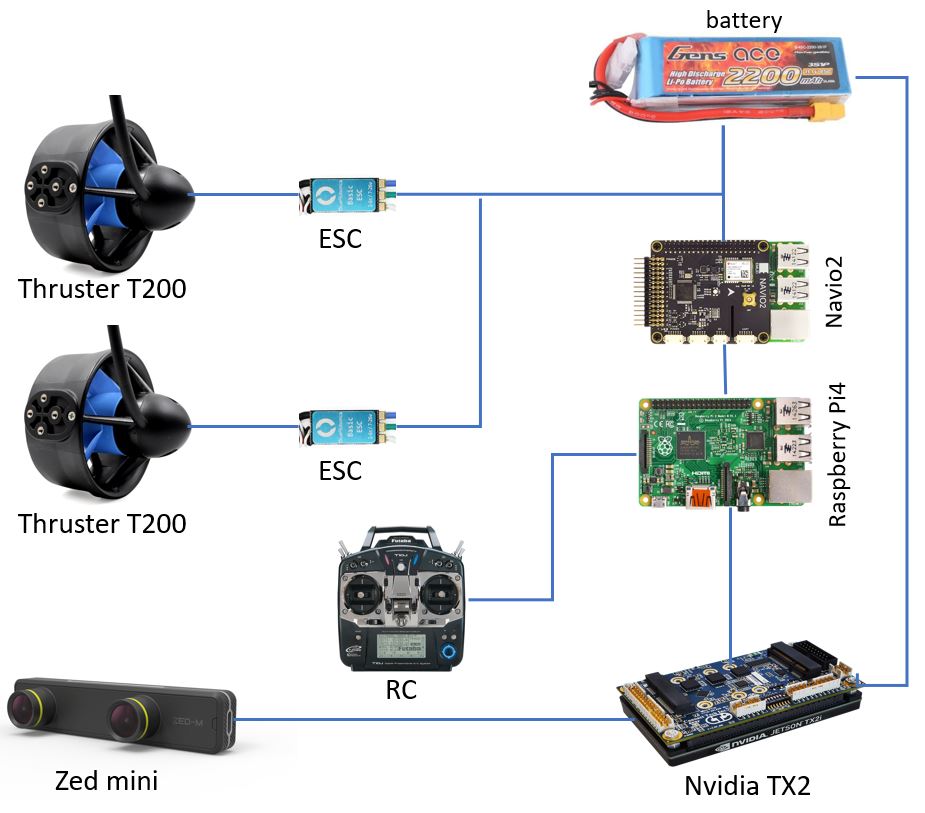}
    \caption{Schematic of sensors, actuators, and electronic components of the marine$\mathcal{X}$}
    \label{fig:electronic}
\end{figure}
\section{Hardware design}

The hardware design plays a key role in the vessel's reliability and performance. A well-engineered hardware design ensures the USV's stability, buoyancy, and hydrodynamic efficiency, enabling smooth and efficient navigation in various marine conditions.
This section will explain the marine$\mathcal{X}$ design process and core hardware components such as sensors, actuators, computing units, CAD models, and fabrication process.

\begin{figure}
\begin{center}    
    \includegraphics[width=0.95\columnwidth]{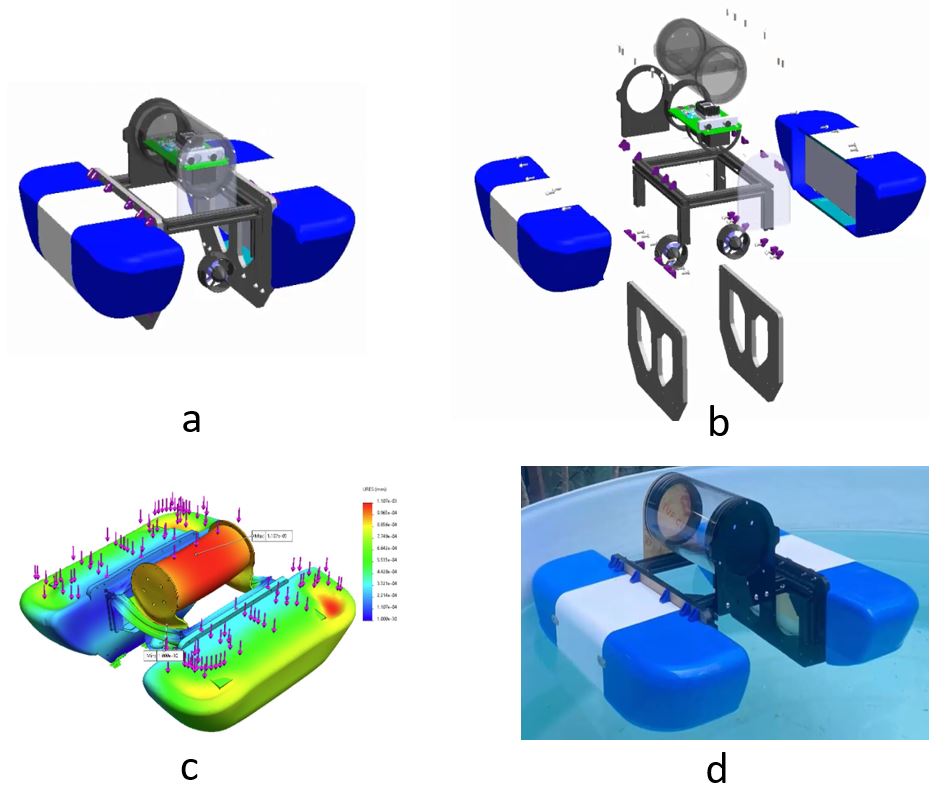}
    \caption{Design and fabrication process of marine$\mathcal{X}$. a) and b) depict the complete CAD design and the shredded view of the components, respectively. c) shows the stress test analysis. d) shows the first buoyancy test of marine$\mathcal{X}$}
    \label{fig:febrication}
    \end{center}
\end{figure}
\subsection{Electronic components}\label{sub:ec}
The main components of the Marine$\mathcal{X}$ are shown in Fig~\ref{fig:electronic}. The actuation system consists of two T200 thrusters by Blue Robotics Inc (\url{https://bluerobotics.com/}). The thrusters include both clockwise and counterclockwise propellers.  The propeller's maximum rotational speed is 4000 rpm and it can produce up to 5.05 kgf thrust. The rpm are controlled using the electronic speed controller (ESC) to ensure the smooth and precise control. The control inputs for autonomous navigation is computed using Navio2 board by  HiPi Industries Inc (\url{https://navio2.hipi.io/}). It is an autopilot header for Raspberry Pi4, it also contains the onboard IMU and GPS sensors.  The Navio2 with the integration of Raspberry Pi4, control the actuators for autonomous navigation.  The USV incorporates the powerful NVIDIA Jetson TX2 computing board. It is responsible for processing advance computer vision algorithms. This module empowers the USV to run the state-of-the-art vision algorithms for accurate object detection tasks on the video stream received from the ZED Mini camera mounted at the front of the USV enclosure. In order to operate the marine$\mathcal{X}$ in a teleportation mode, an RC control and receiver is used developed by Fataba (\url{https://www.futaba.com/}).  The RC receiver is connected to the Raspberry Pi that receive the command from RC control and forward it to the thrusters. To sustain the Marin$\mathcal{X}$ operations, a high-capacity
10000mAh LiPo Battery is employed.  Fig~\ref{fig:electronic} provides a visual representation of the electronic components and connection flow made for the USV.

\subsection{CAD Model and Fabrication}
We choose catamaran design as it evident batter stability over the single hull design for autonomous vessels. The CAD model for the marine$\mathcal{X}$ is shown in Fig.~\ref{fig:febrication}-a and the exploded view of the components is shown in Fig~\ref{fig:febrication}-b. The weight of the the marine$\mathcal{X}$ is 25 kg and it can move with the speed of up-to 4 knots. The hulls are 3D printed and painted with white and blue color. From inside, the hulls are filled with the bouncy foam to provide the batter payload capacity. It can carry payload of up-to 20 kilogram. An aluminium frame is designed using Bosch profiles, it is connected with the hulls via acrylic sheet. Thrusters are also connected with hulls using acrylic sheet, A water tightening enclosure is used to carry all the electronic components such as raspberry pi, TX2, camera and LiPo battery. This enclosure is connected with the frame using acrylic sheet.

\section{Software architecture}
\begin{figure}
    \includegraphics[width=0.95\columnwidth]{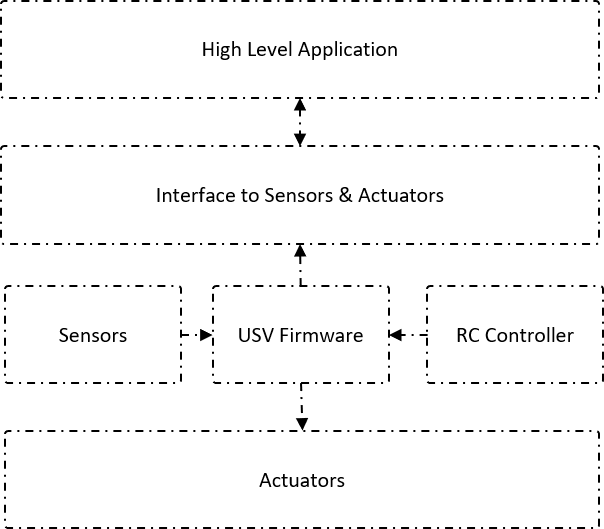}
    \caption{Overview of software architecture of marine$\mathcal{X}$ }
    \label{fig:software}
\end{figure}

The software architecture of the marine$\mathcal{X}$ is shown in Fig.~\ref{fig:software}. The main software system resides in Raspberry Pi, it consists of USV Firmware that provides sensors and actuators interface for the high-level application. The vision-related applications are running on the TX2 that has onboard NVIDIA GPU for fast processing. 
\subsection{USV Firmware}

 USV firmware is a C++ based software application, developed on top of Resbian operating system which is a Linux-based default OS for Raspberry Pi. The communication layer consists of ROS Noetic, it is used to communicate among different hardware components such as sensors are actuators. The USV Firmware module contains the low-level ROS driver to communicate with Sensors (such as GPS and IMU), actuators (such as thrusters), and provide ros-topics to send and receive data from sensors and actuators. Moreover, USV Firmware is also responsible to receive the commands from the remote control and forward them to thrusters for teleportation. 
The high-level application will consist of a navigation algorithm. Any feasible control scheme and navigation algorithm can be implemented as a ROS node in a high-level application module. 

\begin{figure*}
\begin{center}
    \includegraphics[width=0.95\linewidth]{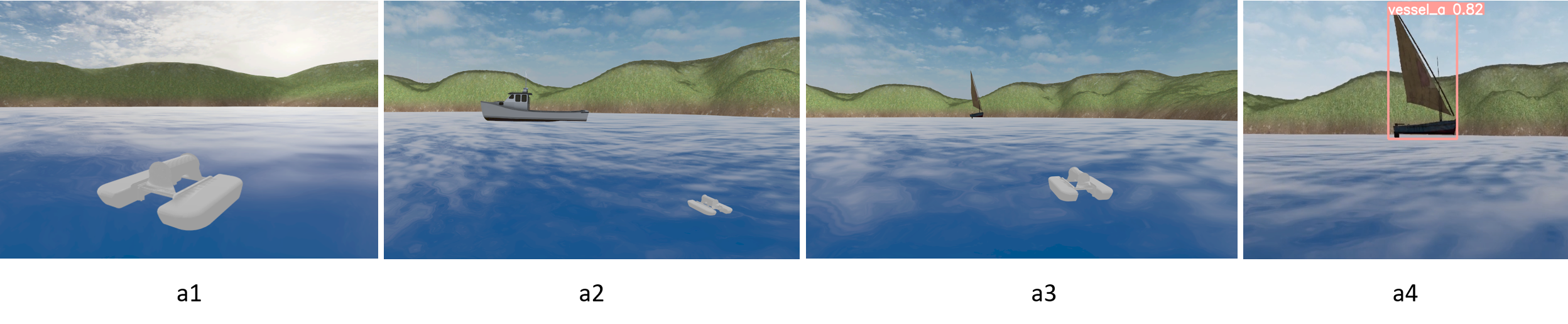}
    \caption{Snapshots of marine$\mathcal{X}$'s in Gazebo simulation. a1-3) show marine$\mathcal{X}$ in different navigation scenes. a4) shows the marine$\mathcal{X}$ camera image after applying the detection algorithm,}
    \label{fig:resultsim}
\end{center}
\end{figure*}
\subsection{ROS-based Gazebo Simulation}
Gazebo is a powerful and widely used open-source robotics simulation software. It enables the realistic emulation of various environmental conditions and vessel dynamics. By creating a virtual environment, we can test and refine control strategies and navigation systems in a safe way. Gazebo simulation model for marine$\mathcal{X}$ is created using a Unified robotic description format (URDF). The hydrodynamic and hydrostatic parameters of marine$\mathcal{X}$ are estimated using ANSYS simulator (\url{https://www.ansys.com}).  

In order to test the navigation algorithms, we used gazebo based open-source marine simulator (\url{https://github.com/osrf/mbzirc.git}) developed by Open Robotics for the MBZIRC Maritime Grand Challenge. It is developed in C++ and Python. It uses ROS2 Galactic for communication among modules. Fig~\ref{fig:resultsim} shows snapshots of the marine$\mathcal{X}$ simulation in MBZIRC simulator. 
The simulation helps in identifying potential issues and improve overall system robustness. It serves as a crucial bridge between theoretical concepts and real-world implementations, ensuring that marine$\mathcal{X}$ can operate effectively and autonomously in diverse maritime applications. The marine$\mathcal{X}$ simulation model could also easily be incorporate with Nav2 (\url{https://navigation.ros.org/}). It is a ROS2-based navigation stack that provides several ready to used path planners for navigation. 

\section{Vision Guided Navigation}\label{sec:vision}
\begin{figure}
    \includegraphics[width=\columnwidth]{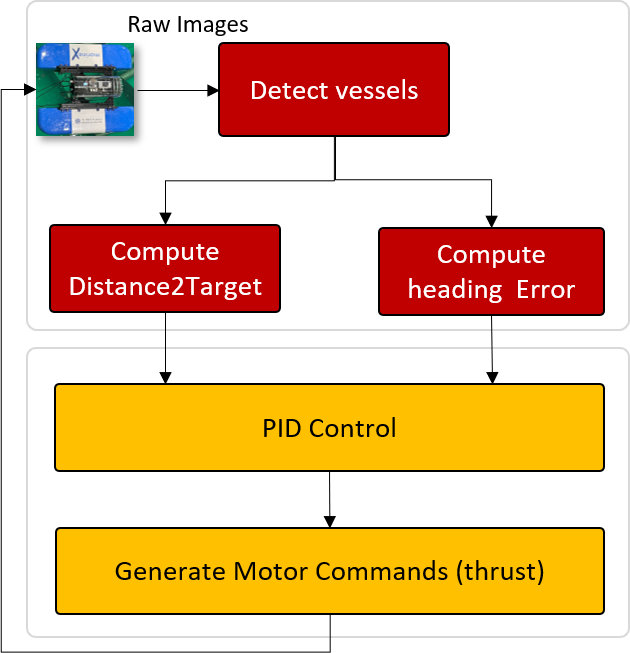}
    \caption{Framework for marine$\mathcal{X}$'s vision guided navigation.}
    \vspace{-2.0em}

    \label{fig:framework}
\end{figure}

Vision-guided navigation consists of applying advanced computer vision techniques to detect the target object and using the control strategies that navigate the USV in a way that it follows (or navigates towards) the detected object. It is an ideal solution for the tasks such as maritime surveillance, environmental monitoring, search and rescue missions, and offshore infrastructure inspection. Fig.~\ref{fig:framework} shows the framework for vision-guided navigation that is implemented on marine$\mathcal{X}$. The raw images captured by the marine$\mathcal{X}'s$ camera are passed to the object detector, SEC.\ref{subsec:vp} explains the vision pipeline for object detection. Based on the detection output, distance and heading error will be computed (details in Sec. ~\ref{subsec:control}). The control law will compute the appropriate control signal that will be used to generate the required thrust. 

\begin{figure}
    \includegraphics[width=0.95\columnwidth]{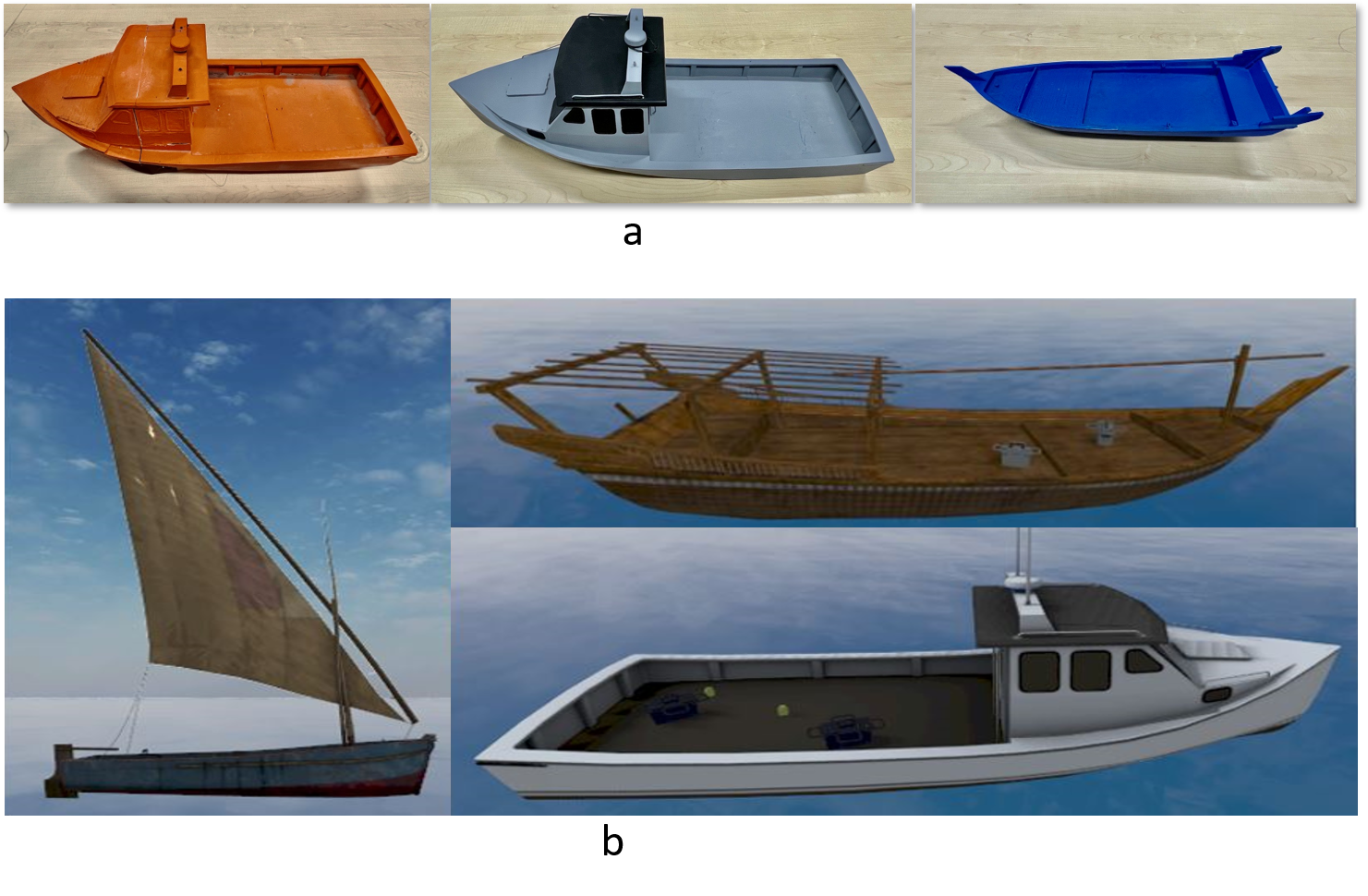}
    \caption{Screenshots of the vessels used to generate the training dataset. Fig-a, show the 3D-printed vessels used for real experiments. Fig-b shows the vessels in the simulation used for gazebo experiments.}
    \label{fig:boats}
\end{figure}
\subsection{Vision Pipeline}\label{subsec:vp}
For object detection, deep learning models are considered highly effective. various deep learning-based object detectors are proposed such as \cite{maskrcnn,fasterrcnn,yolov3}. This work used Yolov5 \cite{yolov5} model for object detection. It is very effective and commonly used for object detection. For real scenario, the model is applied on custom dataset with 3 different 3D printed vessels, shown in Fig.\ref{fig:boats}-a. For simulation, the model is trained on the gazebo models of the vessels as shown in Fig.~\ref{fig:boats}-b, The dataset is recorded by putting the vessels into the marine pool and record the vessels' images using marine$\mathcal{X}$'s onboard camera.

YOLOv5 is a popular deep learning object detection model developed by Ultralytics \cite{yolov5}. It is built using PyTorch (\url{https://pytorch.org/}), a popular deep learning framework, and it follows a one-stage object detection approach, which means it directly predicts bounding boxes and class probabilities for multiple objects in a single forward pass through the neural network. This allows YOLOv5 to achieve faster inference times compared to two-stage object detection models.

The fundamental structure of YOLOv5 relies on a backbone network, often a Convolutional Neural Network (CNN), responsible for extracting essential features from the input image. These extracted features then undergo several subsequent layers, including convolutional, upsampling, and fusion layers, to produce detailed and high-resolution feature maps. In YOLOv5, feature aggregation is enhanced by incorporating the PAN (Path Aggregation Network). This addition allows the model to gather features from various levels of the network, enabling it to effectively capture information at different scales for improved object detection. During the detection process, YOLOv5 follows distinct phases. It performs box and class predictions while also leveraging features from the network's head. By integrating these features from the head, the model can make more precise and comprehensive predictions for object bounding boxes and their corresponding classes. To effectively train the YOLOv5 model, a substantial dataset with labeled examples, including bounding box annotations for the objects of interest, is necessary. The model's training process involves utilizing techniques like backpropagation and gradient descent to iteratively adjust the network parameters and enhance object detection performance. YOLOv5 provides multiple model sizes, such as YOLOv5s, YOLOv5m, YOLOv5l, and YOLOv5x, each varying in depth and computational complexity. These diverse variants offer a balance between inference speed and detection accuracy, empowering users to select the model that best matches their specific needs. 
\begin{figure*}
    \includegraphics[width=\linewidth]{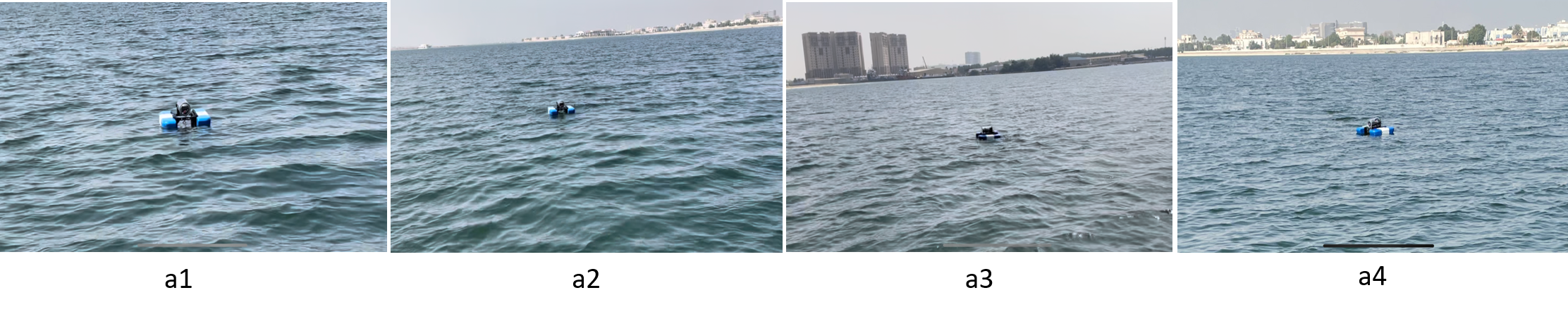}
    \caption{Snapshots of marine$\mathcal{X}$'s open-sea teleportation experiments in the presence of small and large waves. }
    \label{fig:resultsea}
\end{figure*}

\begin{figure*}
    \includegraphics[width=\linewidth]{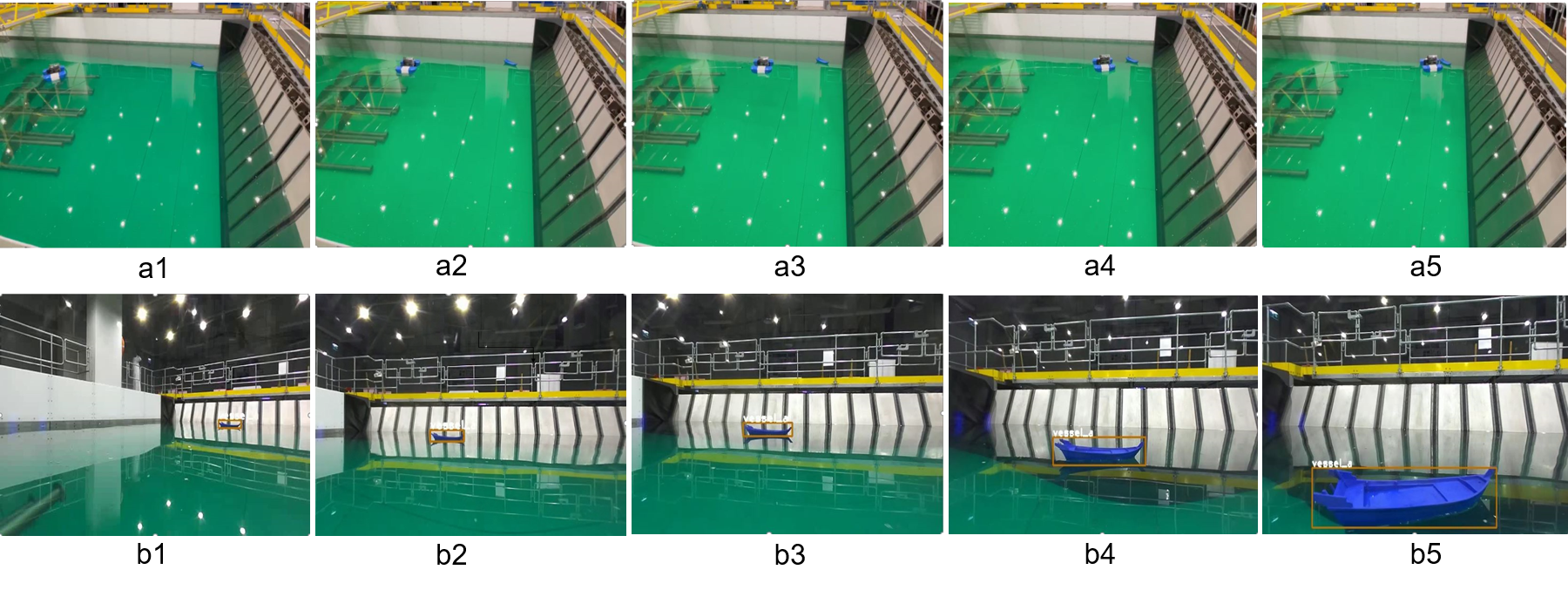}
    \caption{Sequence of snapshots of marine$\mathcal{X}$'s vision guided navigation in the marine pool. Snapshots represented in Fig-a1-5 are captured with the external camera whereas, snapshots in Fig-b1-5 are captured with marine$\mathcal{X}$'s onboard camera.}
          \vspace{-1.0em}

    \label{fig:resultpool}
\end{figure*}

The loss function of YOLOV5 is defined as follows:

\begin{equation} 
L_{yolo} = L_{box} +   L_{cls} + L_{obj} 
\end{equation}

where $L_{box}$, $L_{cls}$, and $L_{obj} $ denotes the bounding box regression, classification, and confidence loss function respectively. 

The bounding box regression is expressed as follows:

\begin{equation}
\begin{aligned}
    L_{box}=\lambda_{coord} \sum_{i=0}^{S^2} \sum_{j=0}^{B} I_{i,j}^{obj} b_j (2-w_i\times h_i)\\ \left[  (x_i-x \wedge_i^j)^2+ (y_i-y \wedge_i^j)^2+(w_i-w \wedge_i^j)^2+(h_i-h \wedge_i^j)^2           \right]
    \end{aligned}
\end{equation}

The classification loss is expressed as follows:

\begin{equation}
L_{cls} = \lambda_{class} \sum_{i=0}^{S^2} \sum_{j=0}^{B} I_{i,j}^{obj} \sum_{c \in \text{classes}} p_i (c) \log (\hat{p}_i (c))
\end{equation}

Similarly, the confidence loss is expressed as below:

\begin{equation}
\begin{aligned}
L_{obj} = \lambda_{noobj} \sum_{i=0}^{S^2} \sum_{j=0}^{B} I_{i,j}^{noobj} (c_i- c\wedge_i)^2\\
+ \lambda_{obj}  \sum_{i=0}^{S^2} \sum_{j=0}^{B} I_{i,j}^{obj} (c_i- c\wedge_i)^2
 \end{aligned}
\end{equation}

where, $\lambda_{coord}$, $\lambda_{class}$, $\lambda_{noobj}$, and $\lambda_{obj}$ are loss coefficient. $w$, $h$,$x$, and $y$ represents width, height, center coordinates, respectively. $\hat{p}_i(c)$ shows the class probability while $c$ shows the class category.

\subsection{Control Strategies for Vision Guided Navigation}\label{subsec:control}

Control strategies for vision-guided navigation of marine$\mathcal{X}$ involve the integration of computer vision algorithms with advanced control techniques to enable precise and autonomous navigation in marine environments. The control scheme is similar to the one used in~\cite{ahmed2023visionbased}. The vision system extracts relevant features from the camera images, such as objects of interest and estimate the marine$\mathcal{X}$'s position and orientation. This information serves as feedback for the control system. To generate control commands for actuators, various control strategies are used, the PID control is a widely used  control technique for such problem.  It is a classical feedback control algorithm  that is designed to regulate a system's behavior by continuously adjusting a control variable based on the difference between the desired setpoint and the current measured value of a process variable. The main goal of PID control is to bring the system's output as close as possible to the desired setpoint while maintaining stability.

The image obtained by the marine$\mathcal{X}$ front camera is passed to the object detector, let $\mathcal{I}$ represent the detector output image. The current position of the target object in $\mathcal{I}$ is represented by $\mathcal{P}_c$ and the desired position of the target object in $\mathcal{I}$ denotes by $\mathcal{P}_d$. The heading error function  $f_e(t)$ is defined as $f_e(t) = \mathcal{P}_d - \mathcal{P}_c$. As an output, it provides the pixel error, that is used to correct the marine$\mathcal{X}$ heading toward the target. To minimize the error function $f_e(t)$, the PID control equation is defined as;
\begin{equation}
    u(t) = K_\textrm{p} f_e(t) + K_\textrm{i} \int_{0}^{t} f_e(t)  \,dt + K_\textrm{d}\frac{d\;f_e(t)}{dt} \label{eq:PID}
\end{equation}

Where $u(t)$ is the control output that will be applied to the marine$\mathcal{X}$ at time $t$. The control signals $u(t)$ is typically used to adjust the marine$\mathcal{X}$ velocity and steering commands to keep the object within the camera's field of view and track its movement accurately. The terms, $K_\textrm{p}$, $ K_\textrm{i}$, and $ K_\textrm{d}$ represent the proportional, integral and derivative gains, respectively. 
In Equ.~\ref{eq:PID}, the integral terms representing the accumulated error over time, it helps to minimize the steady-state errors. The differential term computes the rate of change of error with respect to time. It helps to dampen the system's response and improve stability. The PID control algorithm requires tuning of the the $K_\textrm{p}$, $ K_\textrm{i}$, and $ K_\textrm{d}$, to achieve the desired tracking performance. We tune these parameters in marine pool by manually varying the values of these gains. While navigation, at each iteration, the distance to the target is computed by measuring the depth. The process will continue until the distance to the target vessel is less than given threshold.

\section{Results and Discussions}
We conducted extensive tests to validate the marine$\mathcal{X}$ design and maneuvering capability in simulation (Fig~\ref{fig:resultsim}) and in the  marine pool with no waves and streams. Moreover, to evaluate the operational capability of the marine$\mathcal{X}$ in real scenarios, we also conducted the extensive tests in the open sea environment. Fig~\ref{fig:resultsea} shows snapshots of marine$\mathcal{X}$ operating in open sea. In the teleportation mode, the marine$\mathcal{X}$ shows precise control in the marine pool environment with calm water (in the absence of waves and stream). In the open sea environment it perform well in the presence of small waves and light winds. In the large waves the marine$\mathcal{X}$ navigate accurately, however if we increase the speed, then it jumps on the water surface while navigation. 

In autonomous mode, we tested the vision guided navigation approach (explained in Sec. ~\ref{sec:vision}) in the marine pool, it accurately detect and navigate towards the desired target location. Fig.~\ref{fig:resultpool} show the sequence of snapshots of visual servoing in the marine pool. In open sea the autonomous navigation is possible, however it required to navigate marine$\mathcal{X}$ on a slow speed e.g. 1 knots at maximum. Moreover large waves and strong winds could effect the navigation performance.

\section{Conclusion}
This study introduces marine$\mathcal{X}$, an unmanned surface vessel with a catamaran design. Its robust hardware includes 3D printed hulls, an aluminum frame, and a water-tight enclosure housing the electronics. The vessel's vision-guided navigation algorithm was validated through simulation, marine pool, and open sea tests, exhibiting stable and robust performance across all scenarios. Overall, marine$\mathcal{X}$ showcases the potential of marine robotics and autonomous systems for various maritime applications.

\section*{Acknowledgement}
\noindent This work is supported by the Khalifa University of Science and Technology under Award No. MBZIRC-8434000194, CIRA-2021-085, FSU-2021-019, RC1-2018-KUCARS.

\bibliographystyle{ieeetr}
\bibliography{references}

\smallskip
\end{document}